%% file: iclr2027_conference.tex
\documentclass{article} 
\usepackage{iclr2027_conference,times}

\iclrfinalcopy
\input{math_commands.tex}

\usepackage{hyperref}
\usepackage{url}
\usepackage{graphicx}
\usepackage{amsmath}
\usepackage{booktabs}
\usepackage[table]{xcolor}
\usepackage[most]{tcolorbox}
\usepackage{CJKutf8}
\definecolor{cueblue}{RGB}{235,244,252}
\usepackage{threeparttable}
\usepackage{wrapfig}
\newtcolorbox{promptbox}[1][]{
    breakable,
    colback=gray!4,
    colframe=gray!45,
    boxrule=0.5pt,
    arc=1pt,
    left=4pt,
    right=4pt,
    top=4pt,
    bottom=4pt,
    fontupper=\footnotesize\ttfamily,
    fonttitle=\small\bfseries,
    title=#1
}
\title{CULTURE-CMT: Benchmarking Contextualized Social Media Translation}

\author{
  \textbf{Linjuan Wu\textsuperscript{1,*}},
  \textbf{Dingming Li\textsuperscript{1,2,*}},
  \textbf{Yiwen Qiu\textsuperscript{1,2}},
  \textbf{Xinze Lyu\textsuperscript{1}},
  \textbf{Ye Guo\textsuperscript{1}},
  \textbf{Daoxin Zhang\textsuperscript{1}},
  \\
  \textbf{Zhe Xu\textsuperscript{1}},
  \textbf{Weiming Lu\textsuperscript{2,$\dagger$}}
  \\
  \\
  \textsuperscript{1}Xiaohongshu Inc.
  \quad
  \textsuperscript{2}Zhejiang University
  \\
  \texttt{wulinjuan@xiaohongshu.com}
}

\begin{document}

\maketitle

\begingroup
\renewcommand{\thefootnote}{\fnsymbol{footnote}}
\footnotetext[1]{Equal contribution.}
\footnotetext[2]{Corresponding author.}
\endgroup

\begin{abstract}
Social-media comments are interactions, yet they are often translated and evaluated as isolated text.
This mismatch is especially problematic when meaning depends on surrounding discussion, community-specific slang, or culturally situated expressions.
We introduce \textbf{CULTURE-CMT}, a Chinese--English benchmark for contextualized social-media comment translation that restores the post and conversational context naturally available to users.
Its 572 manually verified examples are organized into \textbf{Slang}, \textbf{Context}, and \textbf{Both}, enabling diagnostic analysis of distinct and interacting sources of translation difficulty.
To support scalable evaluation, we further develop a retrieval-augmented cultural-effectiveness Judger that combines a calibrated scoring model with dynamically retrieved cultural knowledge. CULTURE-CMT reveals several non-trivial patterns.
Providing interaction context improves most model configurations, but does not eliminate culturally ineffective translations.
Examples requiring both contextual grounding and culturally situated language remain substantially more difficult than either source of difficulty alone.
Moreover, larger models are not consistently better, and thinking produces highly model-dependent gains, including occasional degradation.
In contrast, a 27B model adapted on benchmark-disjoint social-media data achieves the lowest observed invalidity rate among the evaluated configurations under thinking inference with context.
Together, these results suggest that realistic social-media translation is not simply a matter of scaling models or reasoning longer, but of learning to translate meaning as it is constructed within evolving online interactions.

\end{abstract}

\begin{figure}[h]
    \centering
    \includegraphics[width=0.85\linewidth]{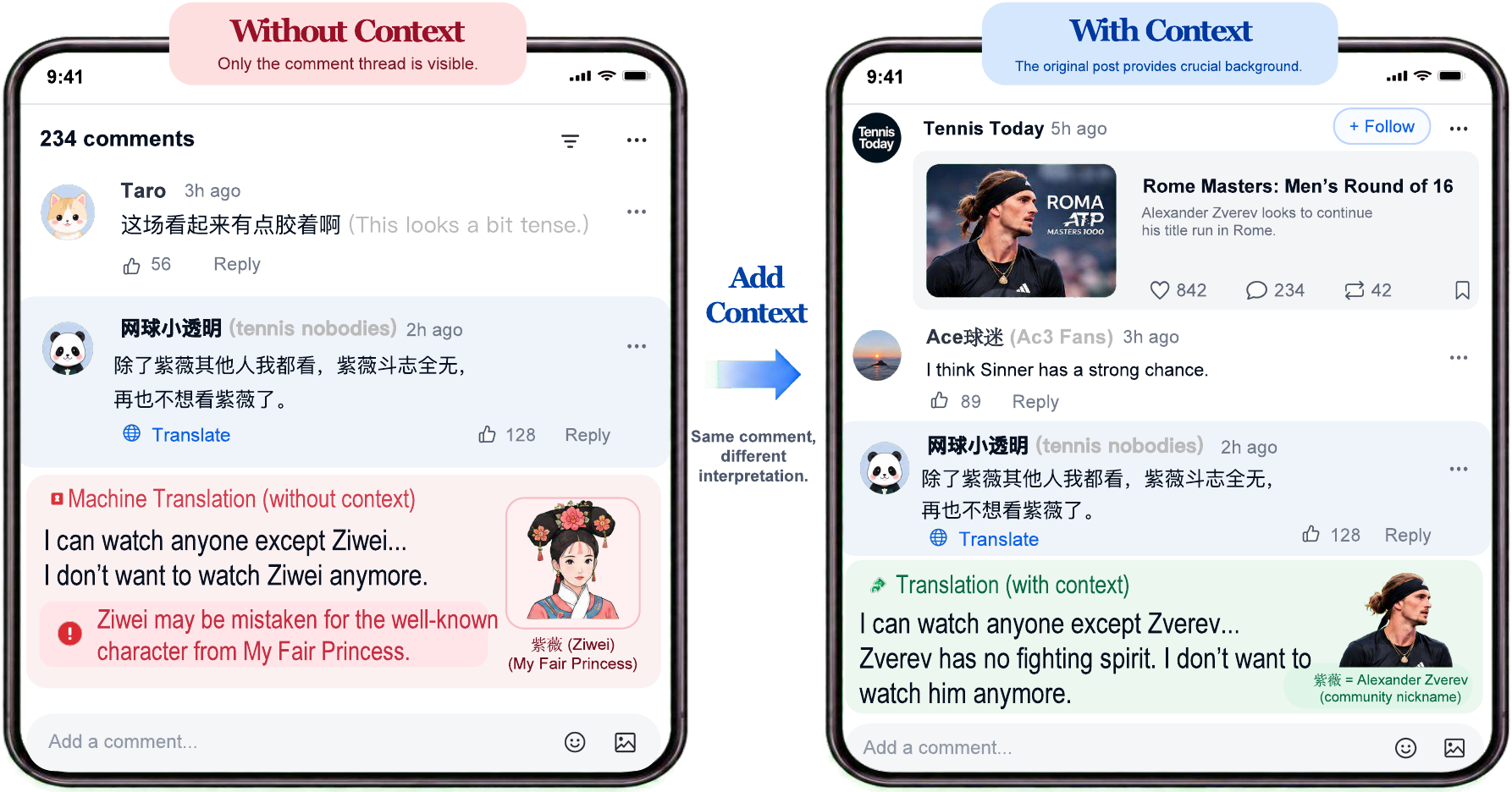}
    \caption{
Interaction context can be necessary to resolve ambiguous social-media expressions and recover the intended translation. The expression ``Ziwei'' is ambiguous in isolation but is correctly resolved as Alexander Zverev when the original post context is available.
}
    \label{fig:motivation}
\end{figure}

\section{Introduction}

Social media platforms are a critical deployment scenario for cross-lingual communication.
Unlike formal genres such as news and encyclopedic writing, user-generated content (UGC) is rich in slang, community-specific expressions, and rapidly evolving non-standard forms~\citep{baldwin2013noisy,michel2018mtnt,sun2024informal}.
Recent benchmarks have begun to study this setting systematically.
CULTURE-MT~\citep{CULTURE-MT} focuses on culturally grounded expressions and adaptation, while CSM-MTBench~\citep{CSM-MTBench} distinguishes longer, information-rich posts from short, reactive snippets.
Related work further shows that slang, neologisms, tone, and other emerging online expressions remain challenging for strong LLMs~\citep{mei2024slang,huang2025buzzwords,slangdit2025}.
Beyond these linguistic challenges, however, \textbf{comments are interactions rather than isolated utterances}: they respond to posts, add information, express stances, and often leave shared information implicit.
Interactional context has been shown to help recover implicit emotion and sociopragmatic meaning~\citep{lee2020eventcomment,zhang2023sociopragmatic}, while context-aware MT benchmarks diagnose discourse-level phenomena~\citep{jwalapuram2020dip,fernandes2023muda}.
Our focus is complementary: we preserve the native post--comment and parent--reply structure of social-media interactions and study it jointly with culturally situated language.

This interaction context can directly change how a comment should be understood and translated.
As shown in Figure~\ref{fig:motivation}, consider the comment
\begin{CJK*}{UTF8}{gbsn}
除了紫薇其他人我都看，紫薇斗志全无
\end{CJK*}
(``I can watch everyone except Ziwei. Ziwei has no fighting spirit.'').
Without the surrounding discussion,
\begin{CJK*}{UTF8}{gbsn}
“紫薇”
\end{CJK*}
(``Ziwei'') is ambiguous and may naturally be associated with the well-known character from the Chinese television drama \textit{My Fair Princess}.
Given the surrounding discussion of the men's round of 16 at the Rome Masters, however, it is correctly resolved as a community-specific nickname for Alexander Zverev.
For platform users, such context is part of the information naturally available when reading the comment, rather than an additional hint.
A small qualitative pilot over 15 randomly sampled comments further suggests that restoring post and conversational context can resolve some translation errors, but does not eliminate culturally ineffective outputs.
Some failures instead involve slang or community-specific expressions, while context can occasionally lead models toward an incorrect interpretation.
This raises a broader question: \textbf{once the original interaction context is restored, what still limits social-media translation?}

To study this question, we introduce \textbf{CULTURE-CMT}, a benchmark for \emph{contextualized machine translation} that evaluates the cultural effectiveness of Chinese--English social-media comment translation.
Each instance preserves the target comment together with its available post title, post content, and parent comment where applicable, covering both top-level comments and replies.
Motivated by our preliminary observations and prior work on slang and emerging internet language~\citep{wuraola2024slang,sun2024informal,huang2025buzzwords}, we categorize examples into \textbf{Slang}, \textbf{Context}, and \textbf{Both} to analyze slang-related difficulty, contextual dependence, and their combination.
We then use CULTURE-CMT as a stress test for three common routes to improving LLM-based translation: \textbf{Scale}, \textbf{Reasoning}, and \textbf{Adaptation}.
This framing is motivated by recent evidence that specialized translation models can rival much larger general-purpose systems~\citep{guo2025redefiningmachinetranslationsocial,cheng2025seedx,zheng2026hymt2familyfastefficient}, while explicit reasoning can benefit challenging interpretative language tasks~\citep{wang2023selfconsistency,slangdit2025}.

Our results reveal a non-trivial pattern.
Larger models are not consistently better, and thinking provides highly model-dependent gains.
A 27B model adapted on benchmark-disjoint social-media data achieves the lowest observed invalidity rate among the evaluated configurations under thinking inference with context.
Beyond aggregate performance, the difficulty breakdown shows that \textbf{Both} remains harder than \textbf{Slang} or \textbf{Context} alone among the analyzed models, highlighting the challenge of combining contextual grounding with culturally situated language.
Together, these results suggest that progress in realistic social-media translation requires more than simply increasing model scale or reasoning compute.

Our contributions are threefold:
\begin{itemize}
    \item We introduce \textbf{CULTURE-CMT}, a Chinese--English social-media comment translation benchmark that preserves the interaction context naturally available to users.

    \item We provide a diagnostic evaluation framework that separates \textbf{Slang}, \textbf{Context}, and \textbf{Both} difficulties, together with a retrieval-augmented cultural-effectiveness Judger validated against human annotations.

    \item We compare \textbf{Scale}, \textbf{Reasoning}, and \textbf{Adaptation}, documenting non-monotonic scaling, model-dependent reasoning gains, and clear benefits from targeted adaptation of the same backbone.
\end{itemize}

\begin{figure}
    \centering
    \includegraphics[width=1\linewidth]{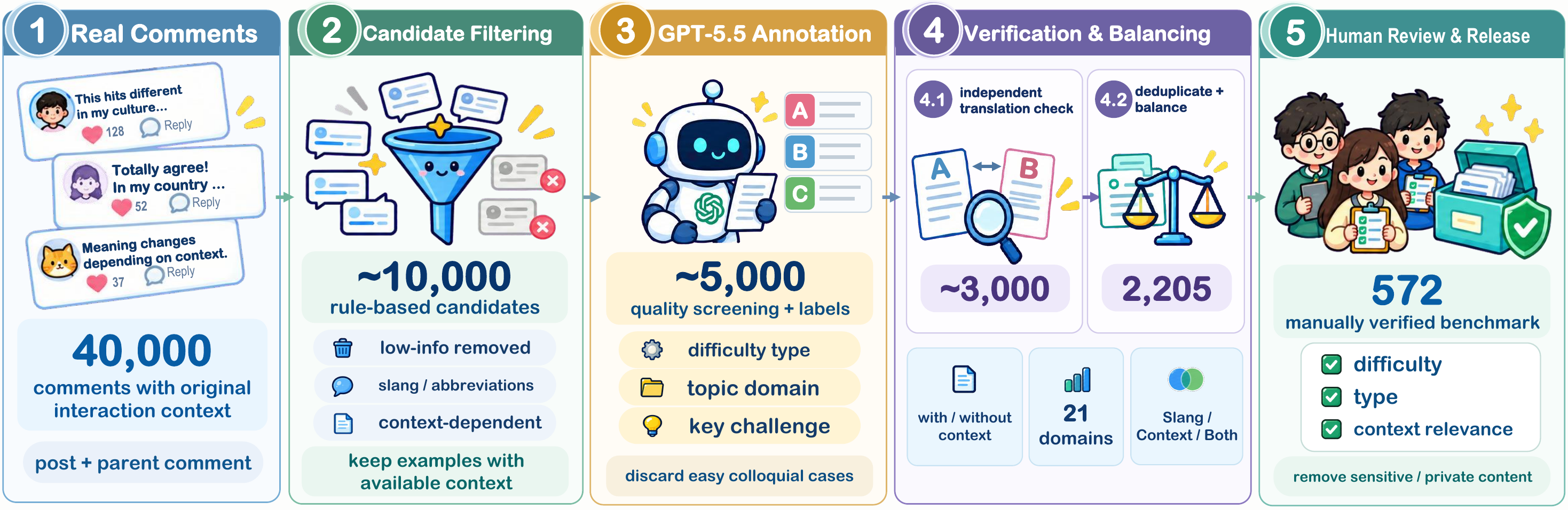}
    \caption{
Construction pipeline of CULTURE-CMT, from 40K contextualized social-media comments to 572 manually verified benchmark examples.
}
    \label{fig:data_pipeline}
\end{figure}
\section{Benchmark Construction}

\subsection{Task Definition}
\label{sec:task_definition}

CULTURE-CMT evaluates Chinese--English social-media comment translation under the interaction context naturally available to users.
Each instance contains a target comment $c$ and its available context, including the post title $t$, post content $p$, and, for replies, the parent comment $r$.
For a top-level comment, $\mathcal{C}=\{t,p\}$; for a reply, $\mathcal{C}=\{t,p,r\}$.

The task is to generate an English translation
\[
y=f(c,\mathcal{C})
\]
that preserves the intended meaning of $c$ in its original interaction context.

For diagnostic analysis, we categorize examples into three difficulty types:
\textbf{Slang}, involving internet slang, abbreviations, community-specific expressions, or other culturally situated usages;
\textbf{Context}, where interpretation depends on the surrounding interaction;
and \textbf{Both}, where both sources of difficulty are present.

\subsection{Construction Pipeline}
\label{sec:construction}

CULTURE-CMT is constructed from naturally occurring social-media comments through a multi-stage pipeline (Figure~\ref{fig:data_pipeline}).
All source data were collected between April 12 and May 12, 2026.
Starting from 40,000 comments with their original interaction context, the pipeline consists of candidate filtering, model-assisted annotation, difficulty verification, deduplication and balancing, and human review.
The resulting benchmark contains 572 manually verified examples.

\paragraph{Candidate Filtering.}
We first apply rule-based quality filters to malformed, overly short or long, template-like, and low-information comments.
We then identify candidates containing either social-media-specific language, such as slang, abbreviations, and community-specific expressions, or potential contextual dependence.
Candidates are required to have their original post context available.
This stage yields approximately 10,000 examples.

\paragraph{Model-assisted Annotation.}
We use GPT-5.5 to perform structured annotation over the candidate pool, including quality screening, difficulty type, topical domain, and the primary translation challenge.
Approximately 5,000 examples are selected for subsequent verification.
The annotation prompt and the 21 domain labels are provided in Appendix~\ref{app:annotation_prompt} and Appendix~\ref{app:domain_labels}, respectively.

\paragraph{Difficulty Verification and Balancing.}
We next verify translation difficulty using independently generated translations.
Candidates are assessed for substantive semantic, cultural, slang-related, or contextual variation across translations.
For \textbf{Context} and \textbf{Both} examples, we additionally compare translations generated with and without interaction context to confirm contextual dependence.
Approximately 3,000 examples remain after verification.

We then apply character-level $n$-gram deduplication and balance the candidate pool across the 21 topical domains and three difficulty types, resulting in 2,205 candidates for human review.
The corresponding prompts are provided in Appendix~\ref{app:translation_generation}, Appendix~\ref{app:difficulty_verification}, and Appendix~\ref{app:context_verification}.

\paragraph{Human Review.}
Three graduate-student reviewers independently examine the 2,205 candidates, covering translation difficulty, difficulty-type validity, and context relevance, with cross-checking across reviewers.
The review also includes screening for sensitive, inappropriate, private, and personally identifiable information.
After human verification, 572 examples remain and constitute CULTURE-CMT.

\begin{figure}[t]
    \centering
    \includegraphics[width=\linewidth]{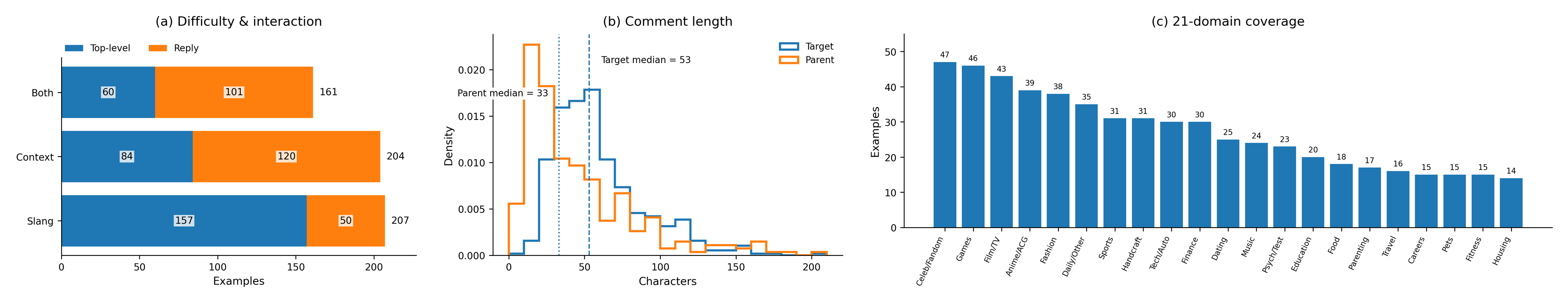}
    \caption{
    Dataset composition by difficulty type, interaction structure, comment length, and topical domain.
    }
    \label{fig:dataset_statistics}
\end{figure}

\subsection{Dataset Statistics}
\label{sec:dataset_statistics}

Figure~\ref{fig:dataset_statistics} summarizes CULTURE-CMT along four dimensions: difficulty type, interaction structure, topical domain, and comment length.

The benchmark contains 207 \textbf{Slang}, 204 \textbf{Context}, and 161 \textbf{Both} examples, together with 301 top-level comments and 271 replies.
Their interaction structures differ noticeably across difficulty types:
\textbf{Slang} is dominated by top-level comments (157/207), whereas replies account for a larger share of \textbf{Context} (120/204) and \textbf{Both} (101/161).
This pattern reflects the greater conversational dependence of the latter two categories.

CULTURE-CMT spans 21 social-media domains.
The largest categories are Celebrities/Fandom (47), Games (46), and Film \& TV (43), while the smallest domain still contains 14 examples, providing broad topical coverage across diverse social-media settings.

Target comments are concise overall but exhibit a long-tailed length distribution, with a median of 53 characters and a mean of 58.9.
Available parent comments are shorter, with a median of 33 characters and a mean of 45.7.
Together, these statistics reflect the short-form yet highly variable nature of social-media interaction.
\section{Evaluation Protocol}
\label{sec:evaluation_protocol}

CULTURE-CMT evaluates whether a translation preserves not only the literal meaning of a social-media comment, but also its culturally situated meaning, affect, and communicative intent.
Following the cultural-effectiveness criterion of CULTURE-MT~\citep{CULTURE-MT}, each translation receives a four-level score
\[
s \in \{0,1,2,3\},
\]
where scores 0--1 indicate culturally ineffective translations and scores 2--3 indicate culturally effective ones.
We adopt the CULTURE-MT scoring rubric, summarized together with our annotation procedure in Appendix~\ref{app:annotation_protocol}.

Our primary metric is the cultural invalidity rate (\textbf{Invalid Rate}),
\[
R_{\mathrm{invalid}}
=
\frac{N_0+N_1}{N},
\]
where $N_k$ is the number of examples receiving score $k$ and $N=572$ is the benchmark size.
Missing translation outputs are assigned score 0.
Lower values indicate better performance.

\begin{figure}[t]
    \centering
    \includegraphics[width=\linewidth]{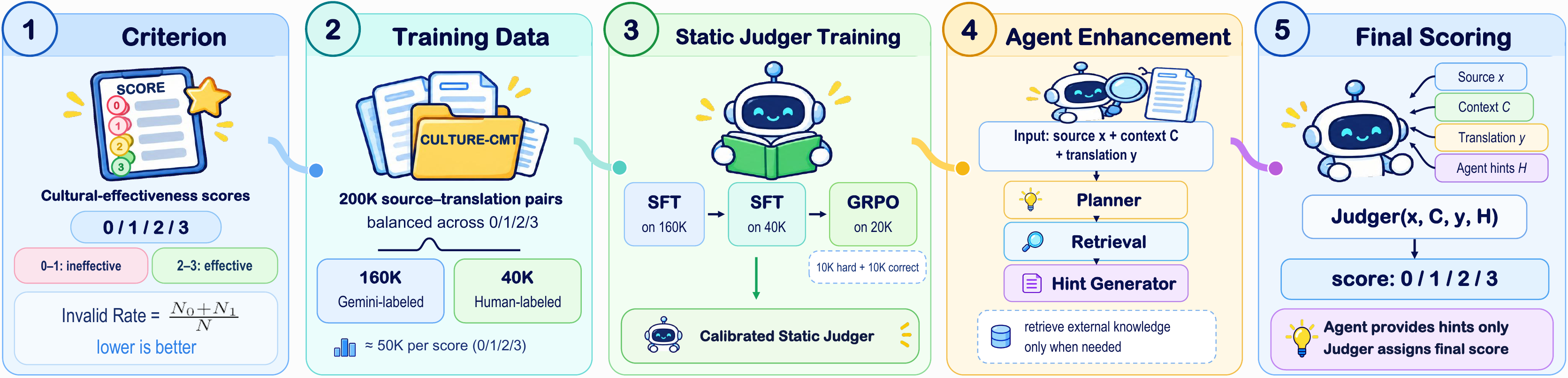}
    \caption{
    Overview of the CULTURE-CMT evaluation framework.
    The cultural-effectiveness Judger is trained through two-stage SFT and GRPO, and is augmented with retrieved cultural knowledge at inference time.
    }
    \label{fig:judger}
\end{figure}

\subsection{Cultural-effectiveness Judger}
\label{sec:judger_training}

To support scalable evaluation, we train a Qwen3.5-27B cultural-effectiveness Judger (Figure~\ref{fig:judger}).
Its training corpus contains 200K source--translation pairs from public social-media data, balanced across the four score levels.
Among them, 160K are labeled by Gemini 3.1 Pro and 40K are manually annotated under the CULTURE-MT criterion~\citep{CULTURE-MT}.
The Judger training data are disjoint from the 40K source pool used to construct CULTURE-CMT.
We additionally reserve development sets of 1,000--2,000 examples during training and a separate 432-pair human-annotated set for final Judger evaluation.
Data separation details are provided in Appendix~\ref{app:data_separation}.

Training proceeds in three stages.
We first perform SFT on the 160K Gemini-labeled examples and then continue SFT on the 40K human-labeled examples.
We subsequently initialize GRPO from the resulting checkpoint using 20K human-labeled examples, comprising 10K examples that remain challenging after SFT and 10K correctly judged examples.
The GRPO reward combines accuracy and format objectives:
correct binary and four-way judgments receive rewards of 1 and 0.5, respectively, while the format reward encourages analysis followed by a final score.
Training uses Relax~\citep{relax2026}; full configurations are provided in Appendix~\ref{app:judger_config}.

The resulting Judger predicts
\[
\hat{s}=f_{\theta}(x,\mathcal{C},y),
\]
where $x$ is the source comment, $\mathcal{C}$ is its interaction context, and $y$ is the candidate translation.

\subsection{Agent-enhanced Judger}
\label{sec:agent_judger}

Social-media evaluation frequently requires knowledge of rapidly evolving slang, memes, fandom terminology, named entities, and platform-specific expressions.
We therefore augment the trained Judger with a retrieval agent that supplies relevant cultural knowledge at inference time.

Given $(x,\mathcal{C},y)$, the evaluation process is
\[
\begin{aligned}
Q &= g_{\mathrm{plan}}(x,\mathcal{C},y),\\
D &= R(Q),\\
H &= g_{\mathrm{hint}}(x,\mathcal{C},y,Q,D),\\
\hat{s} &= f_{\theta}(x,\mathcal{C},y,H),
\end{aligned}
\]
where $g_{\mathrm{plan}}$ identifies knowledge needed for evaluation and generates search queries, $R$ retrieves relevant evidence, and $g_{\mathrm{hint}}$ converts the evidence into concise, translation-specific judging hints.

Retrieved evidence serves as auxiliary knowledge for the calibrated Judger, which produces the final 0--3 cultural-effectiveness score.
This design combines a learned evaluation criterion with dynamically retrieved cultural information, allowing the evaluator to accommodate emerging social-media usages at inference time.
Full agent prompts are provided in Appendix~\ref{app:agent_judger_prompts}.

\section{Experiments}
\label{sec:experiments}
\subsection{Experimental Setup}
\label{sec:exp_setup}

\paragraph{Models.}
We evaluate 33 model configurations spanning multiple families and scales, including Qwen3.5-9B, 27B and 35B-A3B~\citep{qwen35blog}, Qwen3.6-27B\citep{Qwen3.6-27B}, Qwen3.8-27B and Qwen3.8-Max~\citep{qwen38}, GLM-5.2 and GLM-5.3~\citep{glm5team2026glm5vibecodingagentic}, Gemini 3.1 Pro~\citep{google-gemini-3} and Gemini 3.7 Flash~\citep{gemini-37-flash}, Gemma-4-31B and Gemma-4-26B-A4B~\citep{gemmateam2026gemma4technicalreport}, Claude Opus 5~\citep{claude}, GPT-5.6~\citep{gpt-5d6}, and Kimi K3~\citep{kimiteam2026kimik3openfrontier}.
Where both inference modes are available, we evaluate \textit{thinking} and \textit{non-thinking} configurations of the same model.
Cross-model results are interpreted descriptively, as model families differ in architecture and training, while thinking and non-thinking configurations use their native inference settings.

We also evaluate \textbf{Qwen3.6-27B-SFT}, a 27B model adapted for social-media translation using 1.2M training examples strictly disjoint from CULTURE-CMT.
This provides a task-adapted comparison alongside the general-purpose models.
Details of data construction and supervised fine-tuning are provided in Appendix~\ref{app:translation_sft}.

\paragraph{Evaluation.}
We follow the cultural-effectiveness evaluation protocol described in Section~\ref{sec:evaluation_protocol}.
Our primary metric is the proportion of translations receiving scores 0 or 1, denoted as the \textbf{Invalid Rate}; lower values indicate better cultural effectiveness.
We additionally report the proportion of score-3 translations and the mean score.
Judger inference uses greedy decoding (temperature 0).
Contextualized and comment-only translations are evaluated with their corresponding with-context and without-context Judger prompts, so each condition reflects the complete translation-and-evaluation setting available under that context regime.

\subsection{Evaluation of the Cultural-effectiveness Judger}
\label{sec:judger_evaluation}

We assess the Judger against human annotations on translations of final CULTURE-CMT examples.

\paragraph{Data and Metrics.}
For each of the 33 model configurations, we sample 25 source--translation pairs:
5 \textbf{Slang}, 10 \textbf{Context}, and 10 \textbf{Both}, yielding 825 candidates.
After human annotation, we sample by the adjudicated 0--3 labels to construct an approximately score-balanced 432-pair evaluation set.
This set is reserved for final Judger evaluation.
Appendix~\ref{app:annotation_protocol} describes the double-annotation and adjudication procedure.

We report binary accuracy and Cohen's $\kappa$ for the ineffective/effective partition
($\{0,1\}$ versus $\{2,3\}$), together with four-way accuracy and Spearman's $\rho$ over the original scores.
The balanced evaluation set enables comparison across evaluators under comparable score coverage.

\begin{figure}
    \centering
    \includegraphics[width=1\linewidth]{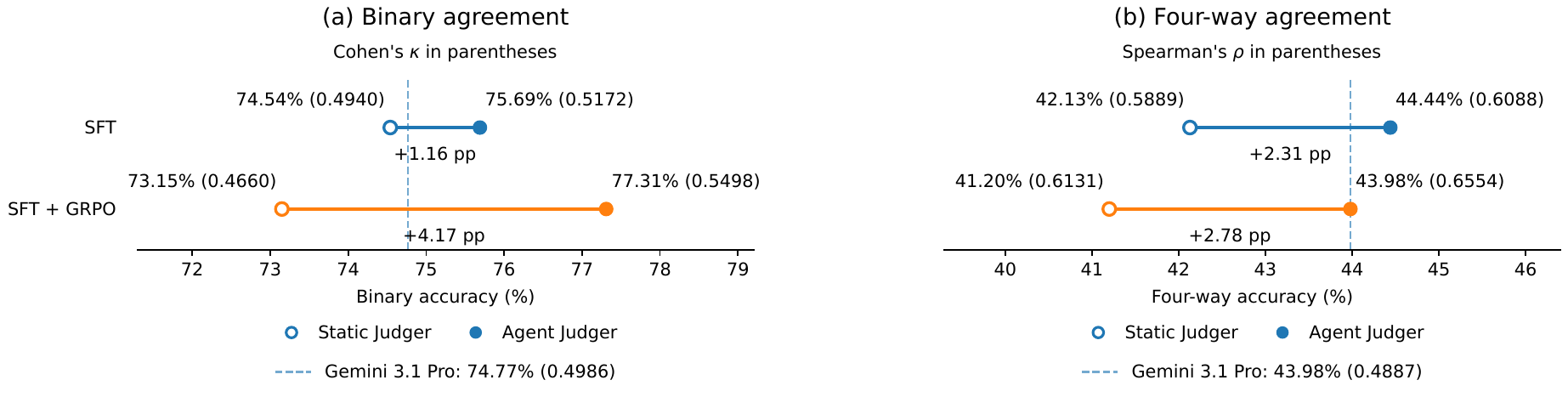}
    \caption{
Agreement with human judgments for static and agent-enhanced Judgers ($N=432$).
(a) Binary accuracy with Cohen's $\kappa$ in parentheses;
(b) four-way accuracy with Spearman's $\rho$ in parentheses. Gains are computed from unrounded accuracies.
}
    \label{fig:judger_retrieval}
\end{figure}
\paragraph{Results.}
The combined agent enhancement increases observed binary accuracy by 1.16 and 4.17 percentage points for the SFT and GRPO-aligned Judgers, respectively, while improving all four reported metrics for both backbones
(Figure~\ref{fig:judger_retrieval}).
GRPO alone improves ordinal agreement
($\rho$: 0.5889 to 0.6131), but reduces binary and four-way accuracy.
The Agent Judger with the GRPO-aligned backbone achieves the highest binary accuracy (77.31\%), $\kappa$ (0.5498), and $\rho$ (0.6554), whereas the SFT-based Agent Judger slightly leads in four-way accuracy.
The main translation experiments use the GRPO-based Agent Judger. Its agreement with humans is assessed on the separate 432-pair evaluation set described above.

\subsection{Overall Results and Difficulty Profiles}
\label{sec:overall_results}

Table~\ref{tab:overall_full} summarizes the complete score distributions
under both contextualized and comment-only settings.
We report the proportions of scores 0--3, the cultural invalidity rate
(\textbf{Invalid}$\downarrow$, i.e., scores 0 and 1), and the mean score.
To quantify the effect of interaction context, we define
\[
\Delta_{\text{Invalid}}
=
R_{\text{Invalid}}^{\text{no-ctx}}
-
R_{\text{Invalid}}^{\text{ctx}},
\qquad
\Delta_{\text{Mean}}
=
\mu^{\text{ctx}}
-
\mu^{\text{no-ctx}},
\]
where positive values indicate better scores in the contextualized condition. Translation and Judger prompts both change between conditions.
Within each inference mode, models are ordered by contextualized
\textbf{Invalid}$\downarrow$.

\begin{table}[t]
\centering

\caption{
Overall CULTURE-CMT results with and without interaction context ($N=572$ in each condition). Missing outputs receive score 0; gains use unrounded values.
Lower \textbf{Invalid} is better; higher \textbf{Mean} is better.
}
\label{tab:overall_full}

\setlength{\tabcolsep}{2.0pt}
\renewcommand{\arraystretch}{0.95}

\resizebox{\textwidth}{!}{
\begin{tabular}{@{}lrrrrrr|rrrrrr|rr@{}}
\toprule
&
\multicolumn{6}{c|}{\textbf{With Context}}
&
\multicolumn{6}{c|}{\textbf{Without Context}}
&
\multicolumn{2}{c}{\textbf{Context Gain}}
\\
\cmidrule(lr){2-7}
\cmidrule(lr){8-13}
\cmidrule(lr){14-15}

\textbf{Model}
& \textbf{0}
& \textbf{1}
& \textbf{2}
& \textbf{3}
& \textbf{Invalid}$\downarrow$
& \textbf{Mean}$\uparrow$
& \textbf{0}
& \textbf{1}
& \textbf{2}
& \textbf{3}
& \textbf{Invalid}$\downarrow$
& \textbf{Mean}$\uparrow$
& $\boldsymbol{\Delta_{\text{Inv}}}\uparrow$
& $\boldsymbol{\Delta_{\text{Mean}}}\uparrow$
\\
\midrule

\rowcolor{gray!15}
\multicolumn{15}{@{}l}{\textbf{Non-thinking inference}} \\
\addlinespace[1pt]

Qwen3.8-Max
& 0.2 & 20.1 & 25.5 & 54.2 & \textbf{20.3} & \textbf{2.337}
& 0.3 & 23.6 & 27.6 & 48.4 & \textbf{24.0} & \textbf{2.241}
& +3.7 & +0.096 \\

GLM-5.2
& 1.6 & 20.1 & 30.6 & 47.7 & 21.7 & 2.245
& 1.0 & 29.0 & 29.7 & 40.2 & 30.1 & 2.091
& \textbf{+8.4} & +0.154 \\

Qwen3.8-27B
& 0.3 & 21.3 & 29.9 & 48.4 & 21.7 & 2.264
& 0.7 & 24.7 & 26.6 & 48.1 & 25.3 & 2.220
& +3.7 & +0.044 \\

Claude Opus 5
& 0.7 & 22.2 & 28.8 & 48.3 & 22.9 & 2.247
& 0.5 & 27.3 & 27.8 & 44.4 & 27.8 & 2.161
& +4.9 & +0.086 \\

GPT-5.6 Terra
& 1.0 & 21.9 & 25.7 & 51.4 & 22.9 & 2.274
& 1.0 & 27.6 & 28.7 & 42.7 & 28.7 & 2.129
& +5.8 & +0.145 \\

Gemma-4-31B
& 1.2 & 22.7 & 26.0 & 50.0 & 24.0 & 2.248
& 1.4 & 29.7 & 30.2 & 38.6 & 31.1 & 2.061
& +7.2 & \textbf{+0.187} \\

\rowcolor{blue!6}
\textbf{Qwen3.6-27B-SFT (ours)}
& 1.2 & 23.6 & 29.7 & 45.5 & 24.8 & 2.194
& 1.2 & 29.4 & 29.9 & 39.5 & 30.6 & 2.077
& +5.8 & +0.117 \\

Qwen3.5-27B
& 1.2 & 24.1 & 30.2 & 44.4 & 25.3 & 2.178
& 1.9 & 31.1 & 30.9 & 36.0 & 33.0 & 2.010
& +7.7 & +0.168 \\

Qwen3.5-397B
& 0.9 & 24.5 & 28.8 & 45.8 & 25.3 & 2.196
& 1.9 & 29.2 & 26.4 & 42.5 & 31.1 & 2.094
& +5.8 & +0.101 \\

Qwen3.5-122B-A10B
& 1.9 & 26.0 & 29.7 & 42.3 & 28.0 & 2.124
& 3.5 & 30.1 & 29.2 & 37.2 & 33.6 & 2.002
& +5.6 & +0.122 \\

Gemma-4-26B-A4B
& 1.0 & 29.0 & 24.7 & 45.3 & 30.1 & 2.142
& 1.9 & 31.8 & 30.4 & 35.8 & 33.7 & 2.002
& +3.7 & +0.140 \\

Qwen3.6-27B
& 2.3 & 32.3 & 28.7 & 36.7 & 34.6 & 1.998
& 3.1 & 33.6 & 26.6 & 36.7 & 36.7 & 1.969
& +2.1 & +0.030 \\

Qwen3.5-35B-A3B
& 3.8 & 31.1 & 29.4 & 35.7 & 35.0 & 1.969
& 5.1 & 35.7 & 29.2 & 30.1 & 40.7 & 1.843
& +5.8 & +0.126 \\

MiniMax M3
& 3.8 & 32.7 & 30.8 & 32.7 & 36.5 & 1.923
& 3.8 & 32.9 & 28.8 & 34.4 & 36.7 & 1.939
& +0.2 & -0.016 \\

Qwen3.5-9B
& 10.7 & 34.1 & 30.4 & 24.8 & 44.8 & 1.694
& 6.6 & 39.7 & 33.6 & 20.1 & 46.3 & 1.671
& +1.6 & +0.023 \\

\midrule

\rowcolor{gray!15}
\multicolumn{15}{@{}l}{\textbf{Thinking inference}} \\
\addlinespace[1pt]

\rowcolor{blue!6}
\textbf{Qwen3.6-27B-SFT (ours)}
& 0.5 & 17.1 & 27.6 & 54.7 & \textbf{17.7} & \textbf{2.365}
& 0.7 & 22.2 & 28.0 & 49.1 & 22.9 & 2.255
& +5.2 & +0.110 \\

GLM-5.2
& 0.5 & 18.0 & 28.3 & 53.1 & 18.5 & 2.341
& 0.2 & 23.3 & 28.3 & 48.3 & 23.4 & 2.247
& +4.9 & +0.094 \\

GLM-5.3
& 0.7 & 17.8 & 26.0 & 55.4 & 18.5 & 2.362
& 0.9 & 19.2 & 24.5 & 55.4 & \textbf{20.1} & \textbf{2.344}
& +1.6 & +0.017 \\

Qwen3.5-27B
& 0.2 & 18.7 & 30.1 & 51.0 & 18.9 & 2.320
& 0.3 & 23.4 & 30.8 & 45.5 & 23.8 & 2.213
& +4.9 & +0.107 \\

Kimi K3
& 0.3 & 19.1 & 26.7 & 53.8 & 19.4 & 2.341
& 0.7 & 21.5 & 28.8 & 49.0 & 22.2 & 2.260
& +2.8 & +0.080 \\

Gemini 3.7 Flash
& 1.0 & 18.7 & 25.0 & 55.2 & 19.8 & 2.344
& 0.5 & 22.9 & 25.5 & 51.0 & 23.4 & 2.271
& +3.7 & +0.073 \\

Gemma-4-31B
& 1.0 & 19.4 & 25.0 & 54.5 & 20.5 & 2.330
& 0.5 & 24.7 & 26.7 & 48.1 & 25.2 & 2.224
& +4.7 & +0.107 \\

GPT-5.6 Terra
& 0.5 & 19.9 & 27.4 & 52.1 & 20.5 & 2.311
& 1.0 & 28.1 & 27.4 & 43.4 & 29.2 & 2.131
& \textbf{+8.7} & \textbf{+0.180} \\

Qwen3.5-122B-A10B
& 0.3 & 20.5 & 30.1 & 49.1 & 20.8 & 2.280
& 0.2 & 26.4 & 33.2 & 40.2 & 26.6 & 2.135
& +5.8 & +0.145 \\

Gemini 3.1 Pro
& 0.0 & 20.8 & 24.3 & 54.9 & 20.8 & 2.341
& 0.0 & 20.8 & 26.7 & 52.4 & 20.8 & 2.316
& +0.0 & +0.024 \\

Qwen3.5-397B
& 0.2 & 21.2 & 30.4 & 48.3 & 21.3 & 2.267
& 0.5 & 22.2 & 31.5 & 45.8 & 22.7 & 2.226
& +1.4 & +0.042 \\

Qwen3.5-35B-A3B
& 1.0 & 21.3 & 32.9 & 44.8 & 22.4 & 2.213
& 1.6 & 26.9 & 32.7 & 38.8 & 28.5 & 2.087
& +6.1 & +0.126 \\

Qwen3.6-27B
& 0.3 & 22.7 & 28.1 & 48.8 & 23.1 & 2.253
& 2.4 & 26.4 & 29.2 & 42.0 & 28.8 & 2.107
& +5.8 & +0.147 \\

Qwen3.8-27B
& 0.2 & 23.1 & 30.2 & 46.5 & 23.3 & 2.231
& 1.2 & 26.9 & 30.6 & 41.3 & 28.1 & 2.119
& +4.9 & +0.112 \\

Gemma-4-26B-A4B
& 0.9 & 23.3 & 26.9 & 49.0 & 24.1 & 2.240
& 1.0 & 27.4 & 28.3 & 43.2 & 28.5 & 2.136
& +4.4 & +0.103 \\

Claude Opus 5
& 1.2 & 23.1 & 26.2 & 49.5 & 24.3 & 2.240
& 1.0 & 24.3 & 29.4 & 45.3 & 25.3 & 2.189
& +1.0 & +0.051 \\

Qwen3.8-Max
& 0.5 & 24.3 & 28.3 & 46.9 & 24.8 & 2.215
& 0.3 & 25.5 & 26.9 & 47.2 & 25.9 & 2.210
& +1.0 & +0.005 \\

Qwen3.5-9B
& 0.5 & 29.2 & 35.8 & 34.4 & 29.7 & 2.042
& 1.4 & 35.5 & 34.8 & 28.3 & 36.9 & 1.900
& +7.2 & +0.142 \\

\bottomrule
\end{tabular}
}
\end{table}

\paragraph{Effect of Interaction Context.}
Most configurations achieve lower Invalid Rates and higher mean scores when interaction context is available, although the magnitude of the improvement varies substantially across models.
For example, non-thinking GLM-5.2 improves from 30.1\% to 21.7\% Invalid, while thinking GPT-5.6 Terra improves by 8.7 percentage points.
By contrast, thinking Gemini 3.1 Pro has the same Invalid Rate in both settings, with only a small increase in mean score.
These results show that interaction context is broadly useful, but models differ considerably in how effectively they exploit it.

\subsection{Difficulty Profiles under Thinking Inference}
\label{sec:difficulty_types}

We next describe difficulty profiles for the evaluated thinking configurations with context.

\begin{wrapfigure}{r}{0.54\linewidth}
    \centering
    \vspace{-1.0em}
    \includegraphics[width=\linewidth]{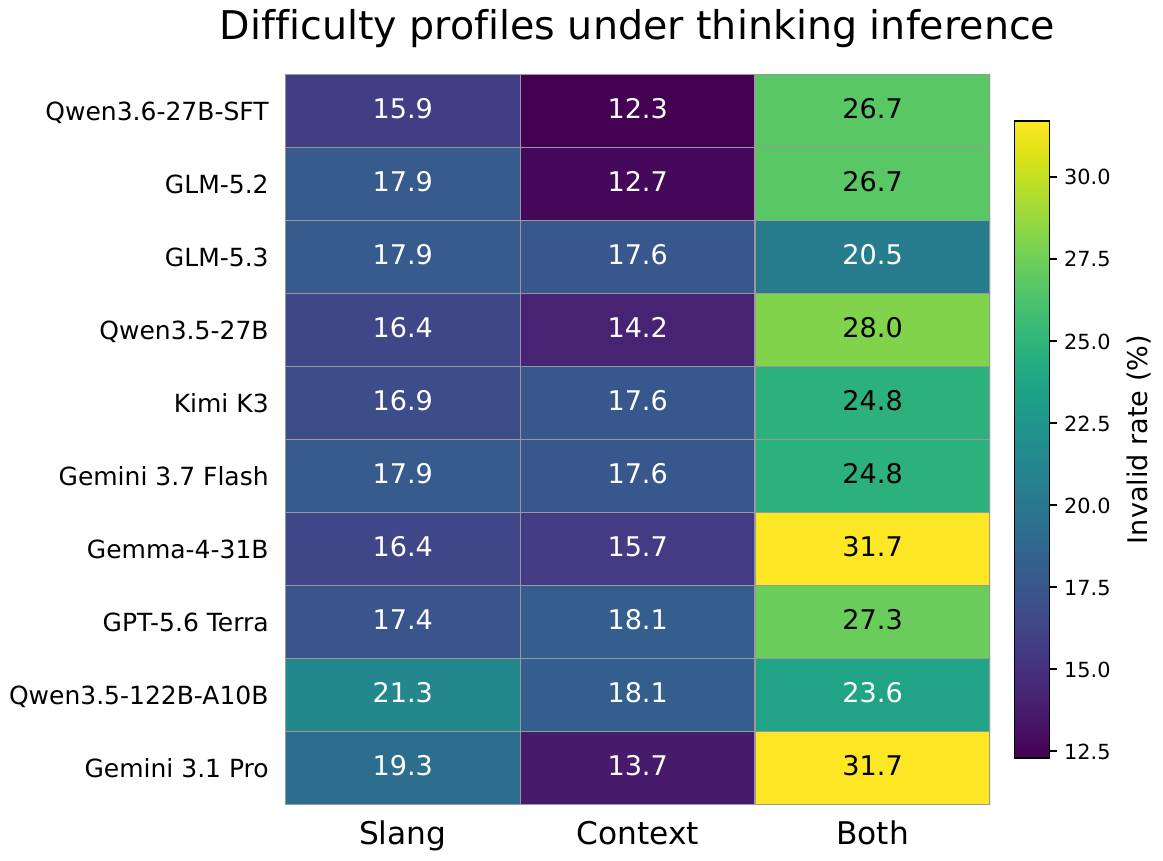}
    \caption{
    Invalid rates (\%) by difficulty under thinking inference with context. Subset sizes are 207 (Slang), 204 (Context), and 161 (Both).
    }
    \label{fig:difficulty_heatmap}
    \vspace{-1.0em}
\end{wrapfigure}

Figure~\ref{fig:difficulty_heatmap} shows Invalid rates on the
\textbf{Slang}, \textbf{Context}, and \textbf{Both} subsets for representative models. Across the models shown, \textbf{Both} is consistently the most challenging subset.
Even systems that perform relatively well on \textbf{Slang} or \textbf{Context} individually often exhibit substantially higher Invalid rates on \textbf{Both}, indicating that the combination of contextual dependence and socially situated language remains difficult.
For example, Qwen3.6-27B-SFT has an Invalid Rate of 26.7\% on Both, compared with 15.9\% on Slang and 12.3\% on Context.
GLM-5.2 shows a similar profile, with 17.9\%, 12.7\%, and 26.7\% on the three subsets, respectively.

The figure also shows that similar overall performance can mask different difficulty profiles.
Some models are relatively stronger on \textbf{Context} but weaker on \textbf{Slang}, while others show the opposite tendency.
For instance, Qwen3.5-27B performs competitively on \textbf{Context} (14.2\%) but remains much weaker on \textbf{Both} (28.0\%).
By contrast, Kimi K3 exhibits a more balanced profile across the three subsets.
The three-way partition thus makes model-specific difficulty profiles visible beyond a single aggregate score.

Our adapted Qwen3.6-27B-SFT has its lowest Invalid Rate on \textbf{Context}, while its higher error rate on \textbf{Both} identifies the combined difficulty as a remaining challenge under thinking inference.

\begin{figure*}[t]
    \centering
    \includegraphics[width=\textwidth]{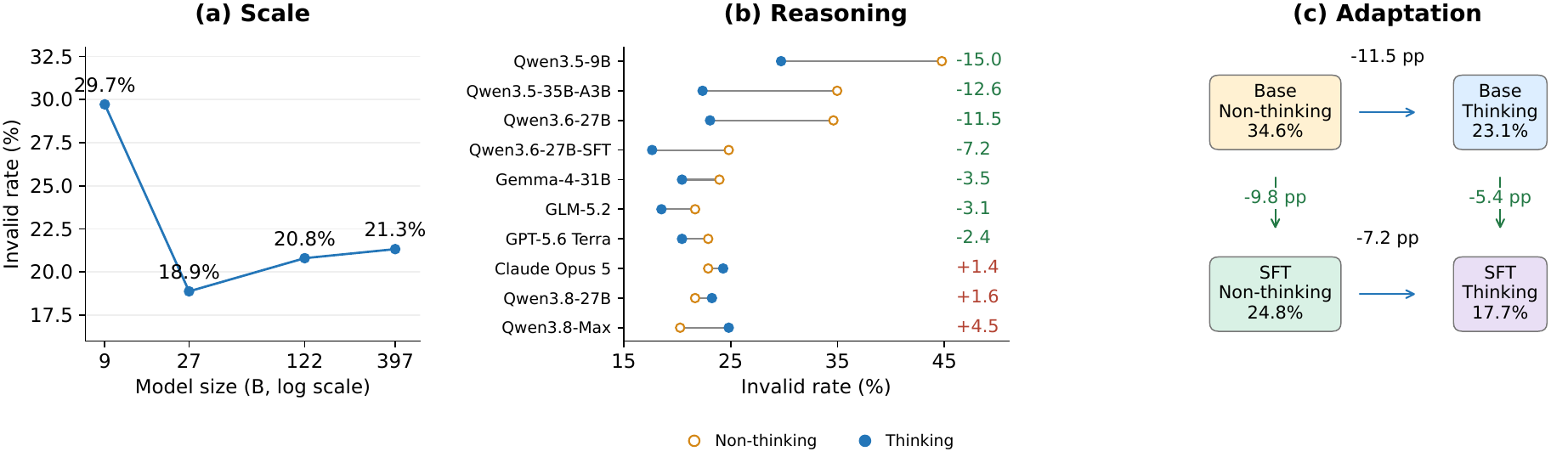}
    \caption{
    Scale, reasoning, and adaptation under the contextualized setting.
    \textbf{(a) Scale:} Invalid Rates across the evaluated Qwen3.5 sizes are non-monotonic.
    \textbf{(b) Reasoning:} thinking/non-thinking differences vary across models.
    \textbf{(c) Adaptation:} SFT of the same 27B backbone lowers observed Invalid Rates in both modes.
    }
    \label{fig:sra}
\end{figure*}

\subsection{Scale, Reasoning, and Adaptation}
\label{sec:sra_analysis}

We next compare three common routes for improving LLM-based social-media translation:
\textbf{Scale}, \textbf{Reasoning}, and \textbf{Adaptation}.
Figure~\ref{fig:sra} summarizes their performance under the contextualized setting.

\paragraph{Scale.}
Within Qwen3.5 under thinking inference, Invalid decreases from 29.7\% at 9B to 18.9\% at 27B, while the 122B-A10B and 397B variants obtain 20.8\% and 21.3\%.
The observed ranking is therefore non-monotonic in total parameter count. For mixture-of-experts models, total and active parameter counts also differ.

\paragraph{Reasoning.}
Thinking inference lowers Invalid by 15.0, 12.6, and 11.5 percentage points for Qwen3.5-9B, Qwen3.5-35B-A3B, and Qwen3.6-27B, respectively.
Claude Opus 5, Qwen3.8-27B, and Qwen3.8-Max instead have higher Invalid Rates under thinking inference.
The benefit of switching to thinking mode therefore varies across models.

\paragraph{Adaptation.}
SFT of the same Qwen3.6-27B backbone lowers Invalid from 34.6\% to 24.8\% without thinking and from 23.1\% to 17.7\% with thinking.
The adapted model achieves the lowest observed contextualized Invalid Rate among the evaluated configurations.

\paragraph{Scope.}
CULTURE-CMT is a targeted stress test constructed from social-media data collected between April 12 and May 12, 2026.
Accordingly, our results characterize the evaluated benchmark and model configurations rather than population-level platform error rates.
The scale and reasoning analyses are descriptive comparisons across existing configurations, while the adaptation comparison captures the full effect of the proposed SFT procedure.

\section{Conclusion}

We introduced \textbf{CULTURE-CMT}, a 572-example benchmark for Chinese--English social-media comment translation that preserves the post and conversational context naturally available to users.
The benchmark supports diagnostic analysis through its \textbf{Slang}, \textbf{Context}, and \textbf{Both} subsets under the cultural-effectiveness rubric inherited from CULTURE-MT.
We further develop a Qwen3.5-27B cultural-effectiveness Judger augmented with retrieved cultural information and evaluate its agreement with human annotations. Across the evaluated configurations, interaction context generally improves cultural effectiveness, while the \textbf{Both} subset remains the most challenging among the models analyzed.
Performance is non-monotonic across model sizes, the benefit of thinking varies substantially by model, and targeted adaptation of the same 27B backbone lowers Invalid Rates in both inference modes.
Overall, these findings highlight the importance of jointly modeling interaction context and culturally situated language, and suggest targeted adaptation as a promising direction for improving social-media translation.

\subsection*{AI Use Statement}
We used generative AI tools to assist with manuscript organization and grammatical editing, as well as figure preparation and consistency checks during revision.
The model-assisted data screening, annotation, translation generation and refinement, and reasoning-trace generation used in this study are described in Section~\ref{sec:construction}, Section~\ref{sec:judger_training}, and Appendix~\ref{app:translation_sft}.
The authors manually reviewed the AI-assisted revisions for correctness and consistency with the reported results and take full responsibility for the final content, including the text, figures, analyses, and conclusions.

\bibliography{iclr2027_conference}
\bibliographystyle{iclr2027_conference}

\appendix
\raggedbottom

\section{Related Work}
\label{sec:related_work}
\subsection{Social Media and Culturally Grounded Translation}

Machine translation for social-media text has long been recognized as challenging because of its noisy, informal, and rapidly evolving nature.
One challenge is robustness to linguistic noise, which MTNT~\citep{michel2018mtnt} studies through naturally occurring Reddit comments.
A complementary challenge is that translations must preserve not only understandable content, but also community-specific meanings, tone, and expressive style.
RedTrans~\citep{guo2025redefiningmachinetranslationsocial} addresses this through specialized training for social-media translation, while CULTURE-MT~\citep{CULTURE-MT} evaluates cultural effectiveness through expression accuracy and cultural adaptability.
CSM-MTBench~\citep{CSM-MTBench} further distinguishes longer posts from short, reactive comments, highlighting difficulties in preserving slang, tone, and style.
Related studies examine slang interpretation and cross-cultural nuances more directly~\citep{sun2024informal,wuraola2024slang}. Studies of emerging slang and internet buzzwords further examine how contextual usage helps interpret unfamiliar expressions~\citep{mei2024slang,huang2025buzzwords}.

CULTURE-CMT complements these efforts by evaluating translation within the original interaction in which a comment occurs.
It retains the corresponding post and parent comment where available, preserving the contextual information naturally accessible to readers.
The \textbf{Slang}, \textbf{Context}, and \textbf{Both} subsets additionally support separate analyses of linguistic-cultural difficulty, contextual dependence, and their combination.

\subsection{Context-aware and Discourse-level Translation}
DiP diagnoses anaphora, lexical consistency, coherence/readability, and discourse connectives in machine translation~\citep{jwalapuram2020dip}.
MuDA uses a data-driven procedure to identify translation phenomena requiring context and supports multilingual discourse evaluation~\citep{fernandes2023muda}.
CULTURE-CMT complements this line of work by retaining explicit post and parent-comment fields and selecting culturally situated comment-translation challenges.
Its contribution is the combination of interaction structure, difficulty-based selection, three diagnostic subsets, and cultural-effectiveness evaluation, rather than the introduction of context into MT evaluation.

\subsection{Scale, Reasoning, and Adaptation for Translation}

Large-scale multilingual training has advanced machine translation~\citep{nllbteam2022nllb}, while specialized systems demonstrate the potential of more compact models.
Seed-X~\citep{cheng2025seedx} develops 7B multilingual translation models, and Hy-MT2~\citep{zheng2026hymt2familyfastefficient} similarly targets efficient translation in real-world settings.
RedTrans~\citep{guo2025redefiningmachinetranslationsocial} explores domain-specific training for social-media translation, while SlangDIT~\citep{slangdit2025} investigates reasoning-based interpretative slang translation.

We bring these directions into a shared evaluation setting to examine \textbf{Scale}, \textbf{Reasoning}, and \textbf{Adaptation}.
Specifically, we compare models across sizes, evaluate thinking and non-thinking modes within the same model where available, and include a social-media translation model fine-tuned on benchmark-disjoint data.
The contribution is a descriptive assessment under a common cultural-effectiveness criterion, rather than a controlled decomposition of the three strategies.
Our model coverage does not include Seed-X or Hy-MT2. Prior CULTURE-MT results place the evaluated Seed-X variants below Qwen3-32B on cultural effectiveness~\citep{CULTURE-MT}, motivating our focus on Qwen-family baselines; this earlier result does not establish their relative performance on CULTURE-CMT. The omission of specialized MT systems limits the scope of the comparison.

\section{Prompts for Benchmark Construction}
\label{app:construction_prompts}

This appendix provides the prompts used in the construction of CULTURE-CMT.
All prompts operate on the original Chinese social-media content.
For clarity, we present the prompts in English while preserving the exact task definitions and output formats used in our pipeline.

\subsection{Model-assisted Candidate Annotation}
\label{app:annotation_prompt}

After rule-based candidate mining, we use an LLM to determine whether each candidate contains a genuine translation challenge and to assign its difficulty type.
The model is given the target comment together with its original interaction context.

\begin{tcolorbox}[
    breakable,
    colback=gray!5,
    colframe=gray!40,
    title=Candidate Annotation Prompt
]

\textbf{System Prompt}

You are an expert reviewer for a challenging social-media machine translation benchmark.
Evaluate each example strictly according to the criteria below and return only valid JSON.

\vspace{0.5em}

\textbf{Input}

Target comment:
\{comment\}

Post title:
\{post\_title\}

Post content:
\{post\_content\}

Parent comment:
\{parent\_comment, if available\}

\vspace{0.5em}

\textbf{Task}

Determine whether the target comment is suitable for a challenging Chinese--English social-media translation benchmark.

An example should be marked as \texttt{pass} only if it contains a genuine translation challenge.
Ordinary conversational expressions that can be translated directly and reliably should be marked as \texttt{reject}.

Assign one of the following difficulty types:

\begin{itemize}
    \item \textbf{Context}: The intended meaning of the target comment cannot be reliably recovered from the comment alone and requires the surrounding post or conversation.

    \item \textbf{Slang}: The target comment contains genuine internet slang, abbreviations, community-specific expressions, culturally situated usages, or other non-standard language that requires non-literal translation or cultural adaptation.

    \item \textbf{Both}: The example contains both substantial contextual dependence and slang/cultural-language difficulty.
\end{itemize}

Do not label an example as \textbf{Context} merely because the context is related to the comment.
The context must be important for determining the intended meaning or an appropriate translation.

Do not label ordinary colloquial expressions as \textbf{Slang}.
The expression should involve genuine internet, community-specific, or culturally situated usage.

Also identify:
\begin{itemize}
    \item the topical domain of the target comment;
    \item the main translation challenge(s).
\end{itemize}

\vspace{0.5em}

\textbf{Output Format}

\begin{verbatim}
{
  "quality": "pass" or "reject",
  "difficulty_type": "Context" or "Slang" or "Both",
  "domain": "...",
  "challenges": ["...", "..."]
}
\end{verbatim}

\end{tcolorbox}

\subsection{Independent Translation Generation}
\label{app:translation_generation}

For candidates that pass the annotation stage, we generate independent translations for difficulty verification.
The translation model receives the target comment and, when required by the experimental condition, its interaction context.

\begin{tcolorbox}[
    breakable,
    colback=gray!5,
    colframe=gray!40,
    title=Translation Generation Prompt
]

\textbf{System Prompt}

Translate the following Chinese social-media comment into natural English.
Preserve its intended meaning, tone, and social-media style.
Use the provided context only to understand the target comment.
Do not translate or summarize the context itself.
Output only the English translation.

\vspace{0.5em}

\textbf{Input}

Post title:
\{post\_title\}

Post content:
\{post\_content\}

Parent comment:
\{parent\_comment, if available\}

Target comment:
\{comment\}

\end{tcolorbox}

For the comment-only condition used in context-dependence verification, the same instruction is used but the post and parent-comment fields are omitted.

\subsection{Translation Difficulty Verification}
\label{app:difficulty_verification}

Model-assisted annotation may retain examples that appear unusual but do not produce substantial translation difficulty.
We therefore generate two translations independently and ask a verifier to determine whether their differences reflect a genuine translation challenge or only stylistic variation.

\begin{tcolorbox}[
    breakable,
    colback=gray!5,
    colframe=gray!40,
    title=Translation Difficulty Verification Prompt
]

\textbf{System Prompt}

You are evaluating whether a social-media comment presents a genuine machine translation challenge.
Return only valid JSON.

\vspace{0.5em}

\textbf{Input}

Target comment:
\{comment\}

Interaction context:
\{context\}

Translation A:
\{translation\_a\}

Translation B:
\{translation\_b\}

\vspace{0.5em}

\textbf{Task}

Compare Translation A and Translation B.

Determine whether their difference reflects a substantive translation challenge in the source comment.

Return \texttt{hard} if the difference is caused by at least one of the following:

\begin{itemize}
    \item different interpretations of the source meaning;
    \item slang, internet language, or community-specific expressions;
    \item culturally grounded or metaphorical language;
    \item implicit or omitted information whose recovery affects the translation;
    \item interaction context that changes the intended interpretation.
\end{itemize}

Return \texttt{style} if both translations convey essentially the same meaning and differ mainly in:

\begin{itemize}
    \item word choice;
    \item sentence structure;
    \item fluency;
    \item degree of formality;
    \item other stylistic preferences.
\end{itemize}

Do not mark an example as \texttt{hard} solely because the two translations use different wording.

\vspace{0.5em}

\textbf{Output Format}

\begin{verbatim}
{
  "reason": "hard" or "style",
  "explanation": "brief reason"
}
\end{verbatim}

\end{tcolorbox}

\subsection{Context-dependence Verification}
\label{app:context_verification}

For candidates labeled as \textbf{Context} or \textbf{Both}, we perform an additional verification step to ensure that the interaction context is genuinely necessary for recovering the intended meaning.

For the same target comment, we generate one translation using the target comment alone and another using the full interaction context.
A verifier then determines whether access to context leads to a substantive improvement or change in interpretation.

\begin{tcolorbox}[
    breakable,
    colback=gray!5,
    colframe=gray!40,
    title=Context-dependence Verification Prompt
]

\textbf{System Prompt}

You are evaluating whether a social-media comment genuinely depends on its interaction context for correct translation.
Return only valid JSON.

\vspace{0.5em}

\textbf{Input}

Target comment:
\{comment\}

Post title:
\{post\_title\}

Post content:
\{post\_content\}

Parent comment:
\{parent\_comment, if available\}

Translation without context:
\{translation\_no\_context\}

Translation with context:
\{translation\_with\_context\}

\vspace{0.5em}

\textbf{Task}

Determine whether the interaction context is necessary to recover the intended meaning of the target comment.

Mark the example as \texttt{context-dependent} if the translation without context:

\begin{itemize}
    \item misinterprets what the comment refers to;
    \item assigns an incorrect meaning to an expression or metaphor;
    \item misses important implicit information required by the target language;
    \item expresses a substantially different stance or communicative intent;
    \item or otherwise fails to recover the intended meaning that becomes clear when the context is provided.
\end{itemize}

Mark the example as \texttt{context-independent} if the comment can be translated reliably without the surrounding context, even if the context makes the interpretation slightly clearer or more specific.

The existence of topical relevance alone is not sufficient for a \texttt{context-dependent} label.
The context must materially affect the interpretation or translation.

\vspace{0.5em}

\textbf{Output Format}

\begin{verbatim}
{
  "label": "context-dependent"
           or "context-independent",
  "explanation": "brief reason"
}
\end{verbatim}

\end{tcolorbox}

\subsection{Domain Labels}
\label{app:domain_labels}

We use 21 topical domains for annotation and balancing:
Games; Anime / ACG; Film and Television; Celebrities / Fandom; Music;
Relationships / Dating; Psychology / Personality Tests; Food; Fashion / Beauty;
Fitness / Exercise; Sports; Travel; Housing / Home; Careers / Job Seeking;
Education / Learning; Parenting / Family; Pets; Drawing / Handcraft;
Technology / Automotive; Finance / Investment; and Daily Life / Other.

\section{Agent-enhanced Judger Prompts}
\label{app:agent_judger_prompts}

This appendix provides the prompts used by the retrieval component of the Agent-enhanced Judger described in Section~\ref{sec:agent_judger}.
The retrieval module supplies auxiliary cultural knowledge when necessary, while the trained cultural-effectiveness Judger retains final scoring authority.
All prompts are presented in English for readability.

\subsection{Search Planning Prompt}
\label{app:search_planning_prompt}

The search planner determines whether external knowledge is required to evaluate a candidate translation and generates a small number of searchable queries when necessary.

\begin{tcolorbox}[
    breakable,
    colback=gray!5,
    colframe=gray!40,
    title=Search Planning Prompt
]

\textbf{System Prompt}

You are a pre-analysis assistant for social-media translation evaluation.

Given a Chinese source text, optional interaction context, and its English translation, determine whether external web search is required to correctly evaluate the translation.

Return JSON only.
Do not use Markdown.

\begin{verbatim}
{
  "needs_search": ["query1", "query2"],
  "reason": "brief explanation"
}
\end{verbatim}

Follow these rules:

\begin{enumerate}
    \item Search only when the input contains knowledge-sensitive content, such as:
    \begin{itemize}
        \item internet memes or emerging slang;
        \item fandom, anime, gaming, or community-specific jargon;
        \item platform-specific expressions;
        \item names of brands, works, people, places, or events;
        \item culturally loaded expressions that may be mistranslated literally;
        \item expressions whose meaning may have changed over time.
    \end{itemize}

    \item Do not search for ordinary words, standard grammar, or expressions whose meaning is already clear from the source and context.

    \item Each query should be short and independently searchable.

    \item Preserve Chinese source expressions in Chinese when they are the object of the search.

    \item If an official English alias is required, use both the Chinese expression and the English alias in the query.

    \item Generate at most four queries.

    \item Return an empty list when no external knowledge is required.
\end{enumerate}

\vspace{0.5em}

\textbf{User Template}

\begin{verbatim}
[Source]
{source}

[Context]
{context}

[Translation]
{translation}
\end{verbatim}

\end{tcolorbox}

\subsection{Hint Generation Prompt}
\label{app:hint_generation_prompt}

The hint generator converts retrieved information into concise, translation-specific judging points.
It does not assign the final cultural-effectiveness score.

\begin{tcolorbox}[
    breakable,
    colback=gray!5,
    colframe=gray!40,
    title=Hint Generation Prompt
]

\textbf{System Prompt}

You are a rigorous but restrained assistant for social-media translation evaluation.

Your task is not to assign a final score.
Instead, provide concise and referenceable judging points for a downstream cultural-effectiveness Judger.

Given the source text, interaction context, candidate translation, search queries, and retrieved evidence, identify only the information that is directly relevant to evaluating the translation.

Return JSON only.
Do not use Markdown.

\begin{verbatim}
{
  "judging_points": [
    "point 1",
    "point 2"
  ],
  "risk_flags": [
    "risk label 1",
    "risk label 2"
  ]
}
\end{verbatim}

Follow these requirements:

\begin{enumerate}
    \item Judging points should explain how the candidate translation should be evaluated.

    \item Relevant aspects may include:
    \begin{itemize}
        \item missing or mistranslated source meaning;
        \item slang, meme, or community-specific meaning;
        \item culturally grounded metaphors or references;
        \item correctness of named entities and official aliases;
        \item preservation of affect, stance, humor, or pragmatic intent;
        \item whether a literal translation loses the intended cultural meaning;
        \item whether the register is appropriate for the original social-media context.
    \end{itemize}

    \item Retrieved evidence is auxiliary information only.
    Ignore evidence that is irrelevant, contradictory, or unreliable.

    \item Do not fabricate cultural knowledge that is not supported by either the source, context, or retrieved evidence.

    \item You may identify potential translation problems, but you must not assign a final score of 0, 1, 2, or 3.

    \item For emerging slang, fandom expressions, special forms of address, memes, or titles of works, explain the intended meaning and the potential risk of literal translation.

    \item Generate at most six judging points.

    \item Keep each judging point concise and directly useful for final evaluation.
\end{enumerate}

\vspace{0.5em}

\textbf{User Template}

\begin{verbatim}
[Source]
{source}

[Context]
{context}

[Translation]
{translation}

[Search Queries]
{queries}

[Search Results]
{search_results}
\end{verbatim}

\end{tcolorbox}

\subsection{Hint Injection Template}
\label{app:hint_injection}

The structured hint object is serialized into an auxiliary block and inserted into the Judger input.
The prompt explicitly instructs the Judger to treat the hints as supplementary evidence rather than authoritative labels.

\begin{tcolorbox}[
    breakable,
    colback=gray!5,
    colframe=gray!40,
    title=Hint Injection Template
]

\begin{verbatim}
Referenceable judging points:

These points are auxiliary information only.
Rely primarily on the source text, interaction context,
and candidate translation.
Do not adopt the hints blindly.

1. {judging_point_1}
2. {judging_point_2}
...
Possible risk flags:
{risk_flag_1}, {risk_flag_2}, ...

Please evaluate the candidate translation using the
four-level cultural-effectiveness criterion.

[Source]
{source}

[Context]
{context}

[Translation]
{translation}
\end{verbatim}

\end{tcolorbox}

\section{Social-Media Translation Fine-Tuning}
\label{app:translation_sft}

\paragraph{Data Collection and Refinement.}
We collect 1.5M Chinese comments from public social-media platforms between April 12 and May 12, 2026, together with their available post and parent-comment context.
Following the data-distillation and refinement approach of CULTURE-MT~\citep{CULTURE-MT}, we use Gemini to generate English translations and a static cultural-effectiveness Judger to assess their quality.
Translations requiring improvement are refined by Gemini using Judger feedback.
After filtering and refinement, we retain 1.2M contextualized source--translation examples, strictly disjoint from CULTURE-CMT.
The translation-generation prompts are provided below.

\begin{promptbox}[Prompt for Contextualized Comment Translation]
You are an expert in social media translation. Please translate the
following content into English. Follow these requirements:

1. Placeholders in the form ``\#placeholder-x\#'' must not be
translated. Preserve each placeholder at the corresponding position
in the translated text.

2. Output only the translated result. Do not provide explanations or
any additional content.

3. Preserve structural tags such as
\texttt{<title></title>},
\texttt{<content></content>}, and
\texttt{<comment></comment>}.
The translated result must follow the same tag structure as the input.

When the sentence to be translated contains an unclear reference or
an ambiguous meaning, use the following post or parent-comment context
to determine its intended meaning:

\{context\}

Return only the final translation and no other content.

Sentence to be translated:

\{input\}

Please provide the translated result:
\end{promptbox}

\paragraph{Thinking and Non-thinking Supervision.}
For each retained sample, GLM-5 generates a concise reasoning trace consisting of \emph{context analysis} and \emph{translation points}.
The context analysis summarizes the information relevant to interpreting the comment, while the translation points identify key expressions and explain their translation considerations.
This task-specific supervision encourages attention to contextual and cultural challenges without unnecessarily long reasoning.
The reasoning-generation prompt is provided below.

\begin{promptbox}[Prompt for Reasoning Trace Generation]
You are a multilingual translation expert for Xiaohongshu-style social
media content.

Given the context, source text, and reference translation, identify the
key translation points that are necessary to produce a translation
like the reference.

First, identify important words or phrases directly from the
\textbf{source text}, and then briefly explain their meanings or the
main considerations for translating them.

Keep the analysis concise and direct.

Example:

Context:
None

Source:
\begin{CJK*}{UTF8}{gbsn}
真的是自嗨的剧情，不好看。 这不是高智商，这是走到哪里都有神助的玛丽苏
\end{CJK*}

Translation:
The plot is truly self-indulgent and unengaging; this isn't a display
of high intelligence but a Mary Sue story where the protagonist is
divinely favored and effortlessly succeeds wherever she goes.

Translation Points:

Pay attention to the following points when translating:

1. ``\begin{CJK*}{UTF8}{gbsn}自嗨\end{CJK*}'' refers to a plot that is immersed in the creator's own
self-satisfaction while ignoring the audience's perspective; it can be
rendered as ``self-indulgent plot.''

2. ``\begin{CJK*}{UTF8}{gbsn}高智商\end{CJK*}'' here refers to either a plot that requires logical
reasoning from the audience or characters whose actions demonstrate
intelligence.

3. ``\begin{CJK*}{UTF8}{gbsn}神助\end{CJK*}'' refers to the protagonist repeatedly receiving convenient
external help whenever difficulties arise.

4. ``\begin{CJK*}{UTF8}{gbsn}玛丽苏\end{CJK*}'' comes from the literary term ``Mary Sue'' and refers to
an unrealistically favored protagonist who seems to have a built-in
advantage or halo.

Context:
\{context\}

Source:
\{content\_zh\}

Translation:
\{trans\_en\}

Translation Points:
\end{promptbox}

We organize the retained data into an equal mixture of thinking and non-thinking supervision, with 600K examples in each subset.
Thinking examples include the generated reasoning trace followed by the final English translation, whereas non-thinking examples retain only the final translation as the supervision target.
Both subsets are combined to train a single model that supports evaluation in either inference mode.

\paragraph{Training Configuration.}
We train \textbf{Qwen3.6-27B-SFT} on the combined 1.2M-example dataset using the \texttt{Relax} framework~\citep{relax2026}.
Training runs for one epoch with a global batch size of 256 and a learning rate of $5 \times 10^{-6}$. Development validation examples are separated from the training collection, as described in Appendix~\ref{app:data_separation}.

\section{Data Separation, Annotation, and Judger Configuration}

\subsection{Data provenance and separation}
\label{app:data_separation}
All source collections in this study were gathered between April 12 and May 12, 2026. The 40K benchmark source comments and the 200K Judger source--translation pairs are disjoint collections gathered during the same period; they do not constitute a temporal train--test split.
The 1.2M retained translation-training examples are disjoint from the final CULTURE-CMT benchmark.
Development validation sets contain 1,000--2,000 examples held out from the corresponding training collections. These development sets are separate from the 432-pair final human evaluation of Judger outputs.
The latter consists of multi-model translations of final benchmark examples and is not used for training, validation during training, checkpoint selection, or evaluator selection.

\begin{table}[htbp]
\centering
\small
\caption{Roles of the data collections. Collection sizes describe the data pools; development validation examples are held out from training.}
\label{tab:data_roles}
\begin{tabular}{@{}p{0.25\linewidth}p{0.16\linewidth}p{0.51\linewidth}@{}}
\toprule
Collection & Size & Role and stated separation \\
\midrule
Benchmark source pool & 40K comments & Source of the 2,205 candidates and final 572 examples; disjoint from the Judger collection. \\
Judger collection & 200K pairs & 160K Gemini-labeled and 40K human-labeled pairs for staged training and held-out development validation. \\
Judger GRPO subset & 20K pairs & Subset of the human-labeled Judger collection, not an additional independent collection. \\
Translation collection & 1.5M collected; 1.2M retained & Translation adaptation; retained examples are disjoint from CULTURE-CMT. \\
Development validation & 1K--2K per set & Held out from training collections; used separately from final evaluation. \\
Final Judger evaluation & 825 candidates; 432 retained & Multi-model benchmark translations, human-labeled and subsampled by score; evaluation only. \\
\bottomrule
\end{tabular}
\end{table}

The static Judger used for translation-data refinement and the agent-enhanced evaluator follow the same cultural-effectiveness criterion. Their shared criterion is part of the training/evaluation design, not evidence of independent human confirmation of the adapted model's ranking.
The reported collection-level separation does not by itself establish that all related posts, threads, or recurring expressions are independent across collections.

\subsection{Human annotation and scoring rubric}
\label{app:annotation_protocol}
Nine Chinese--English bilingual experts annotate translation quality in both the training and evaluation collections.
Each batch is annotated twice. In each round, its examples are divided equally among three experts, so each example receives one label per round, not three votes.
The two rounds use six different experts; assignments are randomized across batches rather than fixed throughout the project.
After both rounds, the remaining three experts adjudicate disagreements and determine the final labels.
The mean binary agreement between the two rounds is 82.46\%, with labels 0--1 mapped to ineffective and labels 2--3 to effective. This is an agreement statistic for the two annotation rounds, not nine-way agreement or four-class exact agreement.

Benchmark candidate review is a separate task: three additional graduate students divide the 2,205 candidates equally, review difficulty and context relevance, and cross-check their work. The resulting 572 examples form the benchmark.
For final Judger evaluation, 25 pairs are sampled from each of 33 model configurations (5 Slang, 10 Context, and 10 Both), giving 825 candidate pairs. Human annotation precedes subsampling by adjudicated score to obtain the approximately balanced 432-pair set.

We reuse the four-level CULTURE-MT rubric without changing its definitions~\citep{CULTURE-MT}. Its two dimensions concern fidelity to meaning and affect, and appropriate cultural interpretation and natural target-language expression. The following is a concise summary, not a new scoring scale:
\begin{itemize}
    \item \textbf{0:} Major failure to convey the intended meaning or cultural intent.
    \item \textbf{1:} Some meaning is understandable, but cultural communication remains ineffective.
    \item \textbf{2:} Communication is generally effective, with residual weaknesses in style or cultural expression.
    \item \textbf{3:} Meaning, affect, and culturally situated intent are conveyed effectively and naturally.
\end{itemize}
The effective/ineffective boundary remains between scores 1 and 2. The Gemini judging setup follows CULTURE-MT's protocol, with Gemini 3.1 Pro used in this study and the context input matched to the evaluation condition.

\clearpage
\subsection{Judger training and inference}
\label{app:judger_config}
All trained Judgers use Qwen3.5-27B. GRPO starts from the Judger checkpoint obtained after the two SFT stages, and uses the Relax framework~\citep{relax2026} with a Megatron training backend and SGLang rollouts.
The Judger's GRPO training reward is
\[
r = r_{\mathrm{accuracy}} + r_{\mathrm{format}}.
\]
The accuracy component rewards agreement with the annotated score: 1 for a correct ineffective/effective decision and 0.5 for a correct four-level score. The format component rewards an analysis followed by the final score.
The configuration below records the supplied GRPO launch settings; it does not describe the separate translation SFT run.

\begin{table}[htbp]
\centering
\small
\caption{Judger GRPO launch configuration. Training sampling and deterministic evaluation use different temperatures.}
\label{tab:judger_config}
\begin{tabular}{@{}p{0.47\linewidth}p{0.47\linewidth}@{}}
\toprule
Setting & Value \\
\midrule
Initialization / reference & Qwen3.5-27B Judger SFT checkpoint \\
Rollout iterations / prompts per batch & 120 / 64 \\
Samples per prompt / global batch size & 4 / 64 \\
Training rollout temperature & 0.9 \\
Maximum generated response length & 2,048 tokens (training and evaluation) \\
Evaluation samples / temperature & 1 per prompt / 0 (greedy) \\
Evaluation / saving interval settings & 10 / 10 \\
Initial evaluation & Skipped by the launch configuration \\
Qwen chat-template thinking flag & Disabled (\texttt{enable\_thinking=false}) \\
Advantage estimator & GRPO \\
Policy clipping (lower / upper) & 0.20 / 0.28 \\
KL coefficient / entropy coefficient & 0.00 / 0.00 \\
Optimizer / learning rate schedule & Adam / constant $10^{-6}$ \\
Weight decay / Adam $\beta_1,\beta_2$ & 0.1 / 0.9, 0.98 \\
Training and rollout allocation & Colocated, 32 GPUs \\
Tensor / pipeline / context parallelism & 4 / 1 / 1 \\
Tokens per GPU / rollout engine size & 8,192 / 4 GPUs \\
SGLang static memory fraction & 0.4 \\
Attention / hidden dropout & 0 / 0 \\
\bottomrule
\end{tabular}
\end{table}

Disabling the chat template's built-in thinking mode is distinct from the task's required analysis-before-score output format.
Final Judger scoring uses temperature 0. The training rollout temperature of 0.9 applies only to generating GRPO samples.
Contextualized translations use the with-context Judger prompt; comment-only translations use the without-context prompt. The agent planning and hint templates are given in Appendix~\ref{app:agent_judger_prompts}.

\end{document}

%% file: math_commands.tex
\usepackage{amsmath,amsfonts,bm}

\def\eqref#1{equation~\ref{#1}}

\def\1{\bm{1}}

\DeclareMathAlphabet{\mathsfit}{\encodingdefault}{\sfdefault}{m}{sl}
\SetMathAlphabet{\mathsfit}{bold}{\encodingdefault}{\sfdefault}{bx}{n}

